\documentclass[a4paper,11pt]{article}
\pdfoutput=1 

\usepackage{jinstpub} 

\title{\boldmath Adaptive Machine Learning for Time-Varying Systems: Low Dimensional Latent Space Tuning}

\author[a]{A. Scheinker}

\affiliation[a]{Los Alamos National Laboratory,\\Los Alamos, NM, USA}
\emailAdd{ascheink@lanl.gov}

\abstract{Machine learning (ML) tools such as encoder-decoder convolutional neural networks (CNN) can represent incredibly complex nonlinear functions which map between combinations of images and scalars. For example, CNNs can be used to map combinations of accelerator parameters and images which are 2D projections of the 6D phase space distributions of charged particle beams as they are transported between various particle accelerator locations. Despite their strengths, applying ML to time-varying systems, or systems with shifting distributions, is an open problem, especially for large systems for which collecting new data for re-training is impractical or interrupts operations. Particle accelerators are one example of large time-varying systems for which collecting detailed training data requires lengthy dedicated beam measurements which may no longer be available during regular operations. We present a recently developed method of adaptive ML for time-varying systems. Our approach is to map very high ($N>100$k) dimensional inputs (a combination of scalar parameters and images) into the low dimensional ($N\approx2$) latent space at the output of the encoder section of an encoder-decoder CNN. We then actively tune the low dimensional latent space-based representation of complex system dynamics by the addition of an adaptively tuned feedback vector directly before the decoder sections builds back up to our image-based high-dimensional phase space density representations. This method allows us to learn correlations within and to quickly tune the characteristics of incredibly high parameter systems and to track their evolution in real time based on feedback without massive new data sets for re-training.}


\begin{document}
\maketitle
\flushbottom

\section{Introduction}
\label{sec:intro}

New advanced particle accelerators are being designed and operated with the goal of finer control over the phase space of the accelerated beams than what has previously been achieved.  At advanced light sources such as the Linac Coherent Light Source (LCLS) upgrade, LCLS-II two color mode operation is being developed in which the energies of two bunches separated by tens of femtoseconds are precisely controlled to generate x-rays with precisely tuned energy differences \cite{ref:LCLS}. At the SwissFEL a two color mode is also being developed and advanced nonlinear compression techniques have enabled the production of attosecond hard x-ray FEL pulses \cite{ref:SwissFEL}. The EuXFEL is utilizing advanced superconducting radio frequency (RF) cavities and low level RF controls to accelerate bunches at MHz repetition rates \cite{ref:EuXFEL}. The plasma wakefield acceleration facility for advanced accelerator experimental tests (FACET) upgrade, FACET-II, is planning on providing custom tailored current profiles with fs-long bunches and hundreds to thousands of kA peak currents \cite{ref:FACET2}. At CERN's Advanced Proton Driven Plasma Wake-field Acceleration Experiment (AWAKE) facility the goal is to use transversely focused high  intensity  ($3\times10^{11}$), high  energy  (400  GeV)  protons  from  CERN's  Super  Proton  Synchrotron (SPS) accelerator to create a 10 meter long plasma and wake-fields into which 18.8 MeV 650 pC electron bunches will be injected and accelerated up to 2 GeV \cite{ref:AWAKE}.

The control of intense charged particle beams is challenging due to complex collective effects such as space charge forces and coherent synchrotron radiation (CSR) which severely distort the 6D phase space $(x,y,z,p_x,p_y,p_z)$ of intense bunches, especially during bunch compression. One of the main challenges of beam control is a lack of detailed real-time phase space diagnostics. To achieve more precise control of intense charged particle beams requires the development of new and higher resolution non-invasive beam diagnostics as well as new advanced adaptive control methods which utilize the data from these diagnostics. The vast majority of online real-time bunch-to-bunch beam measurements provide scalar data such as beam position centroids from beam position monitors or beam averaged charge from current monitors. 

\subsection{Limitations of Machine Learning for Time-Varying Systems}
Machine learning (ML) tools are being developed that can learn representations of complex accelerator dynamics directly from data. ML methods have been utilized to develop surrogate models/virtual diagnostics \cite{ref:FACET_LPS_1,ref:FACET_LPS_2,ref:LCLS2_Injector}, neural networks are being used to represent cost functions or optimal policies in reinforcement learning \cite{ref:deep_Q}, powerful polynomial chaos expansion-based surrogate models have been used for uncertainty quantification \cite{ref:PE}, an interesting analysis of the latent space of neural network-based surrogate models has been used for uncertainty quantification \cite{ref:UQ}, convolutional neural networks have been used for time-series classification and forecasting in accelerators \cite{ref:TS}, neural network-based surrogate models have been used to speed up optimization \cite{ref:MO1,ref:MO2}, Bayessian Gaussian processes utilize learned correlations in data/physics-informed kernels \cite{ref:GP_1,ref:GP_2,ref:GP_3,ref:GP_4,ref:GP_5,ref:GP_6,ref:GP_7,ref:GP_8}, and various ML methods have been used for beam dynamics studies at CERN \cite{ref:ML_CERN1,ref:ML_CERN2,ref:ML_CERN3}.

One limitation of ML methods, and an active area or research in the ML community, is the problem of time-varying systems, known as distribution shift. If a system changes with time then the data that was used to train an ML-based tool will no longer provide an accurate representation of the system of interest, and the accuracy of the ML tool will degrade. Distribution drift is a challenge for all ML methods including neural networks for surrogate models, the use of neural networks to represent cost functions or optimal policies in reinforcement learning, and even for methods such as Gaussian processes which utilize learned correlations in their kernels.

\subsection{Adaptive Feedback for Time-Varying Systems}
In order to handle time-varying systems one must utilize feedback which enables real-time adjustments based on measurements. Powerful model-independent feedback control techniques have been developed to automatically compensate for un-modeled disturbances and time-varying system dynamics for unknown, time-varying, high dimensional systems \cite{ref:ES}. The limitation of most adaptive feedback methods is that they are local in nature and therefore can become stuck in local minima or may converge very slowly if their initial conditions are chosen extremely far away from the optimum in a large parameter space. 

\subsection{Adaptive Machine Learning for Time-Varying Systems}
Efforts have begun to combine the robustness of adaptive feedback with the global representations that can be learned with ML methods to develop adaptive machine learning (AML) for time-varying systems. The first such result combined neural networks and model-independent feedback to provide fast, robust, and automatic control over the energy vs time phase space of electron bunches in the LCLS \cite{ref:ES_ML}. Recently, AML methods have been studied in more generality for adaptive tuning of the inputs and outputs of ML tools such as neural networks for time-varying systems \cite{ref:ES_AML}.

\subsection{Summary of Main Results: Adaptive Latent Space Tuning}
\begin{figure*}[htbp]
\centering 
\includegraphics[width=0.95\textwidth]{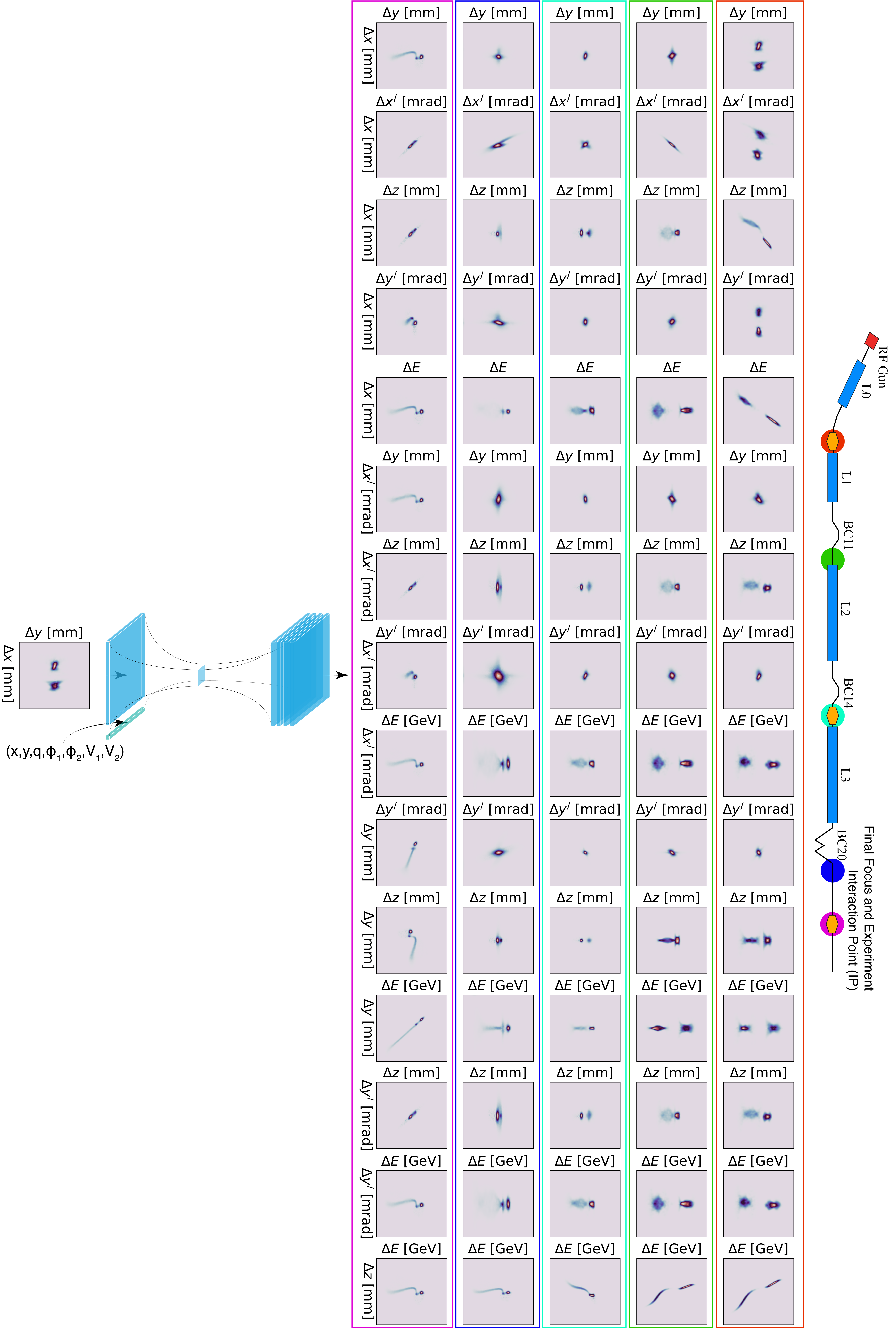}
\caption{\label{fig:75D} One example of a CNN output of 15 phase space projections at the 5 FACET-II locations grouped by colors which are highlighted by the circles on the FACET-II schematic on the right side of the image.}
\end{figure*}

Adaptive latent space tuning maps high dimension inputs down to extremely low dimensional latent space representations of generative encoder-decoder style convolutional neural networks (CNN) \cite{ref:LAML1,ref:LAML2}. In this work we demonstrate an approach in which an encoder-decoder CNN's input is a combination of a 7-dimensional vector representing several particle accelerator components as well as a 128$\times$128 pixel image which is the initial $\rho(x,y)$ phase space density of an electron beam entering the FACET-II accelerator. The overall input dimension of this system, $N_{\mathrm{in}}=16391$, is then mapped down to a latent space dimension as low as $N_L=2$ by the encoder half of the CNN before the generative half builds back up to a 128$\times$128$\times$75 pixel representation of the beam ($N_{\mathrm{out}}\approx$1.2e6) which are all 15 of the unique 2D 128$\times$128 pixel projections of the 6D phase space of the beam at 5 different locations within the FACET-II accelerator, as shown in Figure \ref{fig:75D}. 

By simultaneously predicting 75 phase space projections we are effectively forcing the CNN to satisfy a large number of physics constraints. This low-dimensional latent space representation can be quickly adaptively tuned based only on a comparison of a single one of the 75 output distributions, the position vs energy longitudinal phase space prediction which is available for measurement at many accelerators via transverse deflecting cavities. After the network has been trained to learn the physics constraints within the system and the correlations between various parameters, it functions in an un-supervised fashion with one of the outputs compared to its measurement which then provides feedback to the latent space layer without need for measurements of an input beam distribution or knowledge of accelerator parameters which may drift. Because the network has learned to respect constraints and correlations it is able to predict the entire 6D phase space of the beam at 5 different locations. The feedback compensates for unknown time-varying input beam distributions and accelerator parameters. The low-dimensional latent space representation also informs the placement of limited numbers of diagnostics. The main results are summarized as:
\begin{enumerate}
	\item An encoder-decoder generative convolutional neural network is trained using a supervised learning approach. The CNN takes as input the initial $(x,y)$ input beam distribution and 7 accelerator parameters which are beam $x$ and $y$ offsets, charge, and the amplitudes and phases of the first 2 linear acceleration sections of FACET-II. The CNN pinches down to a low dimensional latent space before the generative half of the CNN outputs all 15 unique projections of the beam's 6D phase space at 5 different accelerator locations as in Figure \ref{fig:75D}.
	\item Once the CNN is trained, we operate in an adaptive un-supervised manner in which we no longer need access to the 7 accelerator parameters or to the initial beam distribution. Our adaptive approach compares the CNN's prediction of the position vs energy $(z,E)$ phase space to transverse deflecting cavity measurements and adaptively tunes the latent space in order to track unknown time-varying beams and accelerator parameters.
	\item The CNN is able to predict all of the 2D phase space projections because it has learned the physics relationships and correlations within the system.
\end{enumerate}

\section{Robust Feedback Control for Particle Accelerators}

Particle accelerators are large systems with thousands of coupled components and complex charged particle dynamics. Accelerators and beams are time varying due to external disturbances, misalignments, and time variation of both the accelerator beam sources as well as of the RF and magnet components such as phase shifts in RF cables. The time variation and complexity of accelerator systems results in time varying input-output relationships, where we refer to inputs as component settings, such as magnet power supply currents or RF accelerating voltages, and outputs as beam characteristics such as beam spill and the beam's 6D phase space distribution. Because of the complexity and variability, the performance of standard model-based controls is limited and drifts with time. Although individual components can be kept at their desired set points very precisely, the overall system dynamics drifts as it depends on the relationships between various components and the beam. Model-independent feedback methods are a major subject of research in control theory where robustness to un-modeled disturbances and changes to system dynamics is an important consideration. For example, consider the scalar linear system
\begin{equation}
	\dot{x}(t) = ax(t) + bu(t), \label{mi_ex1}
\end{equation}
where the values of $a$ and $b$ are unknown. Such a system cannot be stabilized with simple proportional integral derivative (PID)-type feedback, but if the sign of $b$ is known, for example if $b>0$, a stable equilibrium of (\ref{mi_ex1}) can be established at $x=0$ by the following nonlinear controller
\begin{eqnarray}
	u(t) = \theta(t)x(t), \quad \dot{\theta}(t)  = - k x^2(t), \quad k>0. \label{nlc}
\end{eqnarray}
This approach is robust as it does not depend on a detailed knowledge of system dynamics, which is important in practice because all models are only approximations. However, two major limitations of such an approach are: 1). The sign of the unknown term $b$ must be known and cannot be time-varying. 2). The presence of an arbitrarily small un-modeled disturbance in the dynamics (\ref{mi_ex1}) can destabilize the closed loop nonlinear system. Over the last decades, many more sophisticated nonlinear and adaptive feedback control algorithms have been developed which are robust to a wide range of un-modeled disturbances and are applicable to time-varying systems \cite{ref:Khalil,ref:Ioannou1,ref:Ioannou2}. 

A limitation of nonlinear and adaptive control approaches was an inability to handle a sign-changing time varying coefficient $b(t)$ in system (\ref{mi_ex1}) which multiplies the control input $u(t)$, such as $b(t) = \cos(2\pi f t)$ which changes sign repeatedly thereby changing the effect of control input $u(t)$. For particle accelerators such variation is important. For example, consider a state $x(t)$ which describes beam loss in a particle accelerator, whose minimization is desired, which is influenced by a large collection of quadrupole magnets $\mathbf{u}=(u_1,\dots,u_m)$. The effect of a single magnet, $u_m$, depends on the initial beam's phase space as it enters the accelerator from the source and also on the settings of all of the other quadrupole magnets that are upstream, $u_{i<m}$ and changes with time as the upstream magnets are adjusted and as the initial beam conditions change. One day decreasing $x$ may require decreasing the current of magnet $u_m$ and another day it might have to be increased. 

Recently, a nonlinear extremum seeking (ES) feedback control method was developed which could stabilize and minimize the analytically unknown outputs of a wide range of dynamics systems, scalar and vector-valued which can be time-varying, nonlinear and open loop unstable with unknown control directions \cite{ref:ES,ref:ES2}. The ES method is applicable to a wide range of systems of the form
\begin{eqnarray}
	\dot{x} &=& a(t)x(t) + b(t)u(\hat{y}(t)), \quad \dot{\mathbf{x}} = A(t)\mathbf{x}(t) + B(t)\mathbf{u}(\hat{y}(t)), \nonumber \\
	\dot{\mathbf{x}} &=& \mathbf{f}(\mathbf{x}(t),\mathbf{u}(\hat{y}(t)),t), \quad \hat{y}(\mathbf{x},t) = y(\mathbf{x},t) + n(t),
\end{eqnarray}
which include scalar time-varying linear systems, vector-valued time-varying linear systems, and vector-valued nonlinear time-varying systems, where in each case the feedback control $u$ is based only on a noise-corrupted measurement $\hat{y}(t)$ of an analytically unknown cost function $y(\mathbf{x},t)$. For example, a measurable but analytically unknown cost function an be the sum of beam loss along a many kilometer long particle accelerator, which depends on all accelerator parameters and on the initial 6D phase space of the beam being accelerated.

Relative to accelerator applications, the ES method is used to tune groups of accelerator parameters, $\mathbf{p} = \left ( p_1, \dots, p_m \right)$, which may be a subset of the state space $\mathbf{x}$ which describes the overall accelerator dynamics. For example, tuned parameters might include RF cavity amplitude and phase set points as well as magnet power supply voltages or currents. The adaptive ES algorithm dynamically tunes parameters according to
\begin{equation}
	\dot{p}_j = \psi_j \left ( \omega_j t + k \hat{y}(\mathbf{x},t) \right ),
\end{equation}
where $\omega_i$ are distinct dithering frequencies defined as $\omega_i = \omega r_i$ with $r_i \neq r_j$ for $i \neq j$, $k$ is a feedback gain. The $\psi_j$ may be chosen from a large class of functions which may be non-differentiable and not even continuous, such as square waves which are easily implemented in digital systems \cite{ref:ES3}. The only requirements on the $\psi_j$ are that for a given time interval $[0,t]$ they are measurable with respect to the $L^2$ norm and that they are mutually orthogonal in Hilbert space in the weak sense relative to all measurable functions $f(t) \in L^2[0,t]$ in the limit as $\omega \rightarrow \infty$, which can be written as
\begin{eqnarray}
	\lim_{\omega \rightarrow \infty} \int_{0}^{t}\psi_i(\tau)\psi_j(\tau)d\tau &=& 0, \ \forall i \neq j, \quad \lim_{\omega \rightarrow \infty} \int_{0}^{t}\psi_i(\tau)f(\tau)d\tau = 0, \ \forall i, \ \forall f(t) \in L^2[0,t], \nonumber \\
	\lim_{\omega \rightarrow \infty} \int_{0}^{t}\psi^2_i(\tau)f(\tau)d\tau &=& \int_{0}^{t}c_if(\tau)d\tau, \ \forall i, \ \forall f(t) \in L^2[0,t], \ c_i > 0. \nonumber
\end{eqnarray}

One particular implementation of the ES method is especially convenient for particle accelerator applications because the tuning functions $\psi_i$ have analytically guaranteed bounds despite acting on analytically unknown and noisy functions, which guaranteed known update rates and limits on all tuned parameters. This bounded form of ES was first developed for accelerator tuning \cite{ref:ES4}, and then studied for dynamic systems in general \cite{ref:ES5} and takes the simple form
\begin{equation}
	u_i = \sqrt{\alpha_i\omega_i}\cos\left ( \omega_i t + k \hat{y}(\mathbf{x},t) \right).
\end{equation}
The utility of this approach is clearly demonstrated by considering a system of the form
\begin{equation}
	\dot{\mathbf{x}} = \mathbf{f}(\mathbf{x}(t),\mathbf{u}(\hat{y}(t)),t), \quad \dot{p}_i = \sqrt{\alpha\omega_i}\cos\left ( \omega_i t + k \hat{y}(\mathbf{x},t) \right), \label{ESp}
\end{equation}
which results in average dynamics that minimize the noise-corrupted unknown function $y(\mathbf{x},t)$:
\begin{equation}
	\dot{\bar{p}}_i = -\frac{k\alpha}{2}\frac{\partial y(\mathbf{x},t)}{\partial p_i}.
\end{equation}
\section{Adaptive Machine Learning: Latent Space Tuning}
\label{sec:AML}

Convolutional neural networks (CNN) are some of the most powerful machine learning (ML) tools which regularly outperform all other methods in complex high-dimensional tasks such as image classification where, for example, using 100$\times$100 pixel photos results in a $10^4$ dimensional input space \cite{ref:krizhevsky2012imagenet}. The main challenge in utilizing ML approaches for particle accelerator beams is the fact that accelerator components and beam initial conditions change unpredictably with time degrading the accuracy of any data-based ML approach to virtual diagnostics. 

\begin{figure*}[htbp]
\centering 
\includegraphics[width=0.6\textwidth]{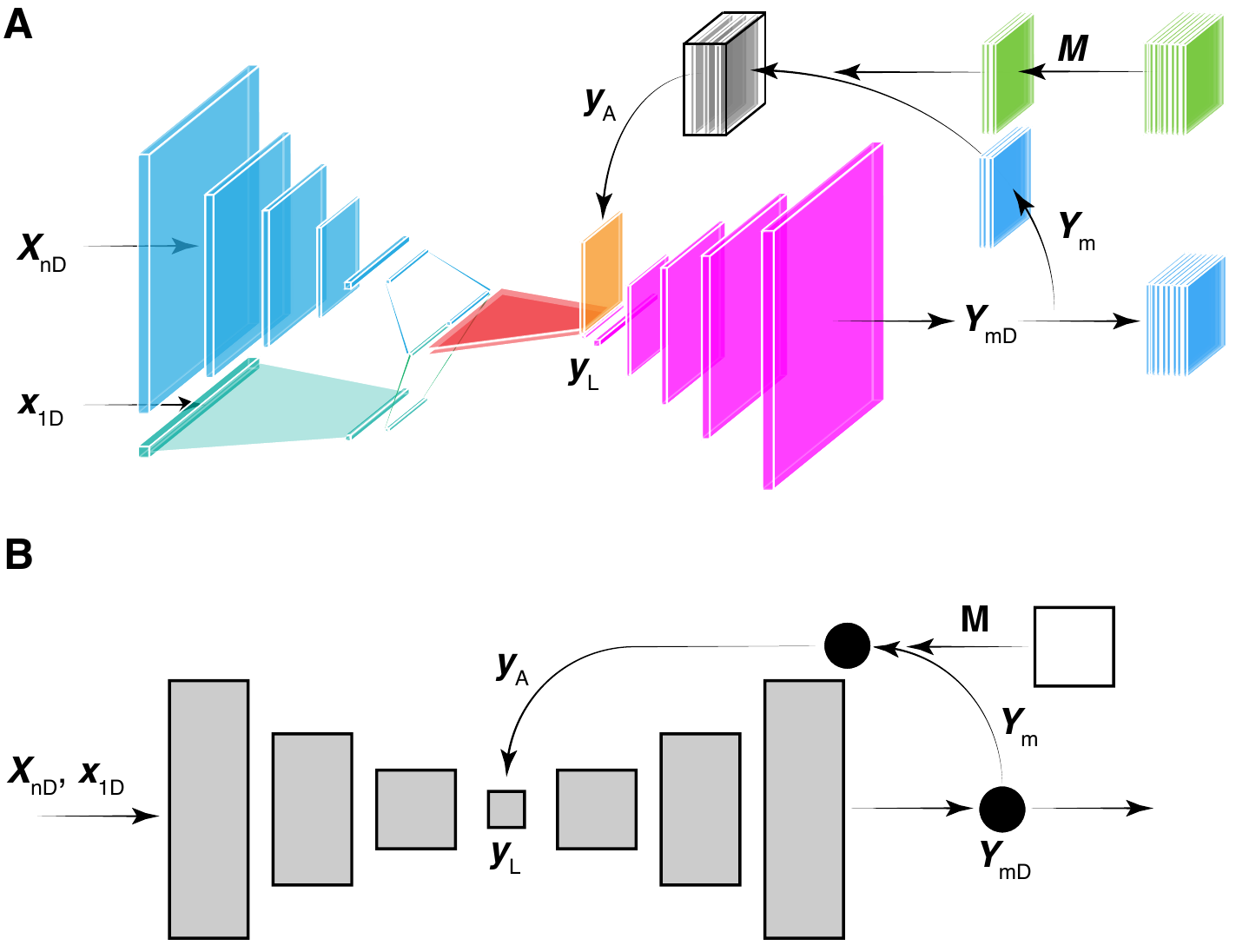}
\caption{\label{fig:aml} An encoder-decoder AML setup is shown in (A) where the inputs are stacks of images $\mathbf{X}_{nD}$, vectors of parameter values $\mathbf{x}_{1D}$, and an adaptively tuned latent space input vector $\mathbf{y}_{A}$, while the output is another stack of images $\mathbf{Y}_{mD}$. A simplified diagram of the same setup is shown in (B). For adaptive tuning, a measurement of the network's output, $\mathbf{Y}_m$, is compared to an available measurement $\mathbf{M}$, based on which the adaptive update $\mathbf{y}_{A}$ is created to adjust the latent space representation $\mathbf{y}_{L}$. }
\end{figure*}

A new adaptive machine learning (AML) approach for time varying systems has recently been proposed in which model-independent adaptive feedback is used to tune the low dimensional latent space of encoder-decoder architecture convolutional neural networks (CNN) \cite{ref:LAML1,ref:LAML2}. The encoder half of the CNN can use both images (initial beam distributions) and scalars (accelerator magnet and RF parameter settings) as inputs with the scalars being concated together with the information from the images once the images have been scaled down using 2D convolutional layers, flattened, and passed through dense fully connected layers. 

The low dimensional latent space at the bottleneck of the encoder CNN is a general nonlinear transformation in which a trained CNN has found an incredibly efficient basis with which to represent beam properties, which may outperform purely linear dimensionality reduction techniques such as principal component analysis \cite{ref:PCA}. The CNN's decoder half can be used to generated predictions of beam phase space distributions as a function of the low-dimensional latent space whose parameters are easily adjustable due to the much lower dimension of the problem which may be orders of magnitude smaller than the many thousands of dimensions associated with large input images.

\begin{figure*}[htbp]
\centering 
\includegraphics[width=1.0\textwidth]{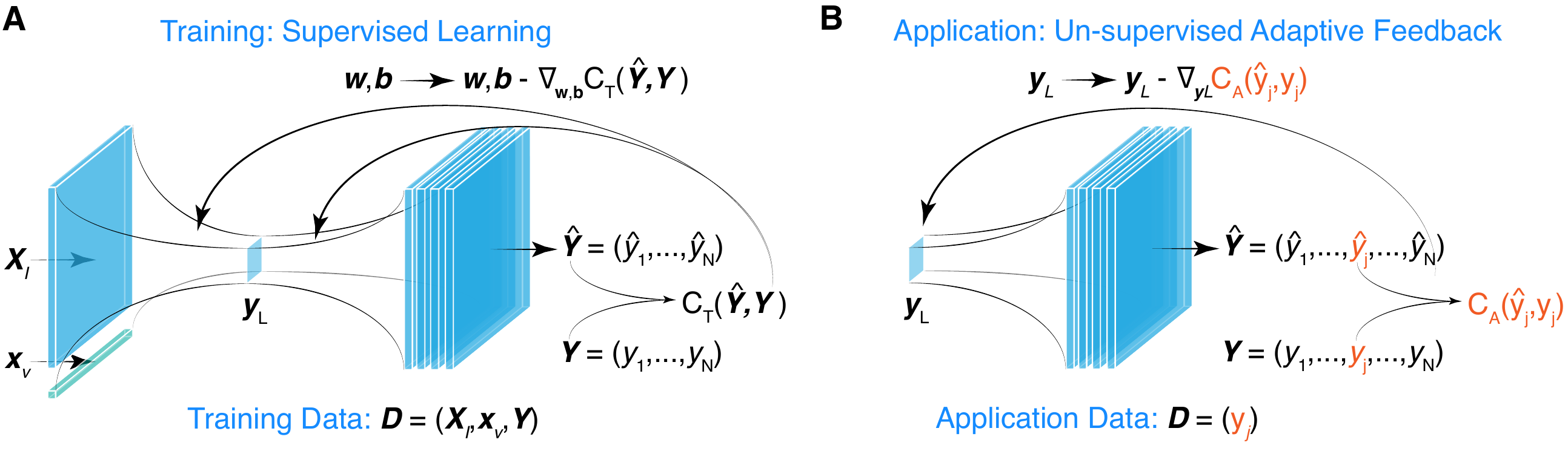}
\caption{\label{fig:L_aml} An overview of the adaptive machine learning latent space tuning approach. }
\end{figure*}

An overview of the adaptive latent space approach is shown in Figure \ref{fig:L_aml}. During the initial training phase a supervised learning approach is taken in which the network uses input-output pairs $(\mathbf{X}_I,\mathbf{x}_v,\mathbf{Y})$, where $\mathbf{X}_I$ may be collections of images which represent 2D projections of a particle accelerator beam's 6D phase space and the $\mathbf{x}_v$ are vectors of parameters such as magnet or RF system settings. The CNN's prediction $\hat{Y}$ of phase space projections is then compared to their true values $\mathbf{Y}$ so that a loss function $C_T(\hat{\mathbf{Y}},\mathbf{Y})$ is minimized by adjusting the CNN weights and biases $(\mathbf{w},\mathbf{b})$ in a supervised training approach as shown in Figure \ref{fig:L_aml}A.  

Once the network has been trained within its weights and activation functions are embedded the relationships and dependencies between the various components of the large output $\hat{Y}$. The trained network is then utilized to make phase space predictions by looking at only those outputs which can be compared to measurements in real time, $\hat{y}_j$, to calculate some cost function $C_{A}(\hat{y}_j,y_j)$ which guides the adaptive tuning of an input to the latent space $\mathbf{y}_{L}$ according to the right hand side of Equation \ref{ESp}, as shown in Figure \ref{fig:L_aml}B.

This approach has two major strengths. The first is that we are now tuning an incredibly high dimensional system within a much smaller space which greatly speeds up the convergence of any optimization or search algorithm. The second main benefit is that once we have learned the generative half of the CNN which is capable of generating complex high dimensional outputs as a function of a low dimensional latent space vector $\mathbf{y}_L$ we no longer need access to the original inputs that were used to train the network. This second feature is especially important for large complex systems, such as particle accelerators, where continuous operations makes it impossible to take detailed beam measurements without interrupting normal operations and therefore is only done during rare dedicated beam development times. The resulting algorithm is robust to uncertain time varying input beam distributions and accelerator parameters as they are assumed to be unknown and the application takes place as an adaptive un-supervised feedback approach which is only allowed to have access to measured quantities which may be a very small subset of all of the CNN's predictions. We demonstrate the approach on two very different particle accelerators: the HiRES compact ultrafast electron diffraction accelerator and the FACET-II plasma wakefield accelerator.

\section{Application for Particle Accelerator Diagnostics}
\label{sec:part}

\begin{figure*}[!tbh]
    \centering
    \includegraphics[width=0.9\textwidth]{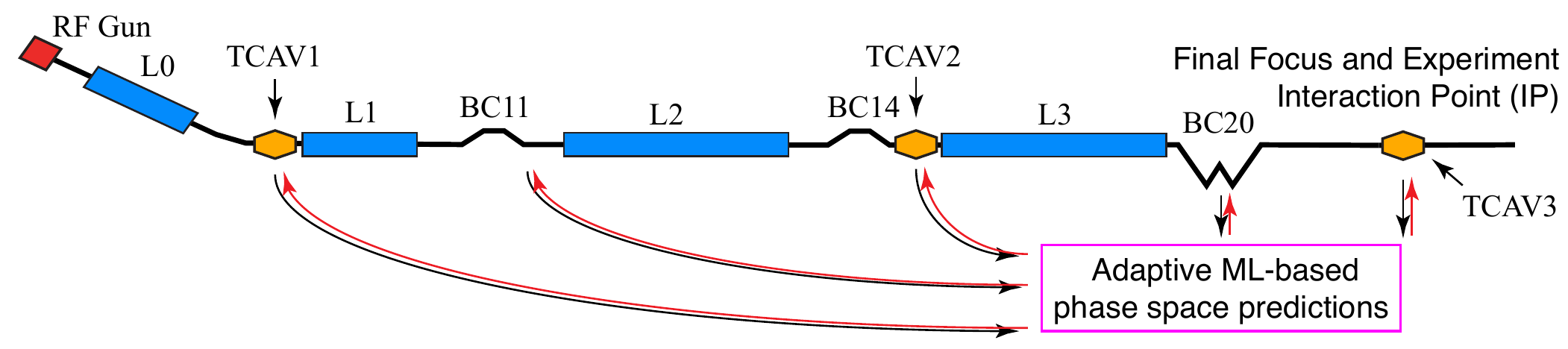}
    \caption{A simplified schematic of the FACET-II beamline is shown where $L0$-$L3$ are linac sections and $BC11$-$BC20$ are bunch compressors. Multiple TCAV locations are shown in yellow, as well as a energy spread spectrum measurement location within bunch compressor $BC20$. The AML approach discussed in this paper can be trained by and provide non-invasive estimates of the beam's phase space at the TCAV locations as implied by the black arrows which represent data that may be used as inputs to the CNN and red arrows that represent possible CNN-based virtual diagnostics once the network has been trained.}
    \label{fig:FACET2}
\end{figure*}

One of the most advanced and detailed diagnostics is a setup using a transverse deflecting RF cavity (TCAV) together with a dipole magnet which separates a beam horizontally as a function of time while separating it as a function of energy vertically resulting in a 2D projection of the beam's $(z,E)$ longitudinal phase space (LPS) with fs resolution \cite{ref:TCAV}. Accelerators such as the LCLS, have a TCAV directly after the undulator so that once the FEL lasing process has completed the remaining electron beam can be separated from the x-rays and the beam's LPS can be measured without disrupting FEL operations. Therefore at the LCLS the TCAV is an always present real-time non-invasive LPS diagnostics. Machines, such as the EuXFEL have their TCAV near the machine's beam source, so it can be used to get an estimate of the LPS characteristics of the beam entering the accelerator. Machines such as FACET-II might have multiple TCAVs, as shown in Figure \ref{fig:FACET2}, which can be used to periodically check the LPS properties of the beams throughout the accelerator.

The LPS of beams has a strong influence over their characteristics in terms of FEL lasing, bunch compression, and wakefield acceleration and therefore many recent efforts have been made for machine learning (ML)-based virtual LPS predictions to complement TCAV-based measurements by providing LPS estimates where TCAVs are not available or by providing LPS estimates without the use of destructive TCAVs. Tools for virtual phase space estimates have been developed which use model-independent adaptive feedback to adaptively tune physics models based on non-invasive beam measurements such as energy spread spectra \cite{ref:ES_FACET}. Once an accurate estimate of a beam's LPS is obtained, whether by using a TCAV-based measurement or a virtual diagnostic, powerful adaptive feedback methods can be used to automatically control the LPS distribution \cite{ref:ES_FACET_2}, including combinations of ML and adaptive feedback for LPS control which is both global and robust to time-variation of accelerator and beam parameters \cite{ref:ES_ML}. 

A latent space tuning AML appraoch is developed for predicting all of the 2D projections of the beams entire 6D phase space given by the 15 distributions:
\begin{eqnarray}
	&& \rho_{12}(x,x'), \quad \rho_{13}(x,y), \ \ \ \quad \rho_{14}(x,y'), \quad \rho_{15}(x,z), \quad \rho_{16}(x,E), \nonumber \\
	&& \rho_{23}(x',y), \quad \rho_{24}(x',y'), \quad \rho_{25}(x',z), \quad \rho_{26}(x',E), \quad \rho_{34}(y,y'), \nonumber \\
	&& \rho_{35}(y,z), \ \ \quad \rho_{36}(y,E), \quad \rho_{45}(y',z), \quad \rho_{46}(y',E), \quad \rho_{56}(z,E), \label{2D_proj}
\end{eqnarray}
where we are now using the coordinates
\begin{equation}
	\mathbf{p} = \left ( p_1, \dots, p_6 \right ) = (x,x',y,y',z,E), \quad x' = \frac{p_x}{p_z}, \quad y' = \frac{p_y}{p_z},
\end{equation}
and the distributions are defined as
\begin{equation}
	\rho_{ij}(p_i,p_j) = \iiiint\rho(\mathbf{p})dp_{i_1}dp_{i_2}dp_{i_3}dp_{i_4}, \quad i_1,i_2,i_3,i_4 \neq i,j. 
\end{equation}
It is usually difficult or impossible to measure the transverse phase space distributions $\rho_{12}(x,x')$ and $\rho_{34}(y,y')$ of the beam quickly and without interrupting operations. Our latent space approach provides an estimate of all 15 2D phase space projections in (\ref{2D_proj}) by comparing the CNN's predictions of the LPS $\rho_{56}(z,E)$ to available TCAV measurements and when possible also projecting $\rho_{16}(x,E)$ distributions down to 1D energy spread spectra when such measurements are available, as was done in \cite{ref:ES_FACET} to adaptive tune an online model to track the $\rho_{56}(z,E)$ phase space non-invasively. This approach relies on the fact that the encoder-decoder CNN has learned correlations between the various phase space projections for a given accelerator lattice, as demonstrated below. 

By training the CNN over a wide range of beam and parameter characteristics, and by forcing the CNN to satisfy the many (15) physics constraints of the 2D phase space projections, the encoder-decoder did learn enough of the underlying beam physics so that it could be used in situations where the input beam distribution is no longer available, but only output beam TCAV-based measurements and possibly energy spread spectra measurements are available. Adaptively tuning the latent space allows us to use the encoder-decoder CNN without needing access to the original input parameter and distribution measurements with which the CNN was trained. This is a major advance for ML-based virtual diagnostics because outside of dedicated beam development times it is often very difficult to gather detailed measurements on input beam distributions in a non-invasive manner that does not interrupt accelerator operations.

\begin{figure*}[!tbh]
    \centering
    \includegraphics[width=1.0\textwidth]{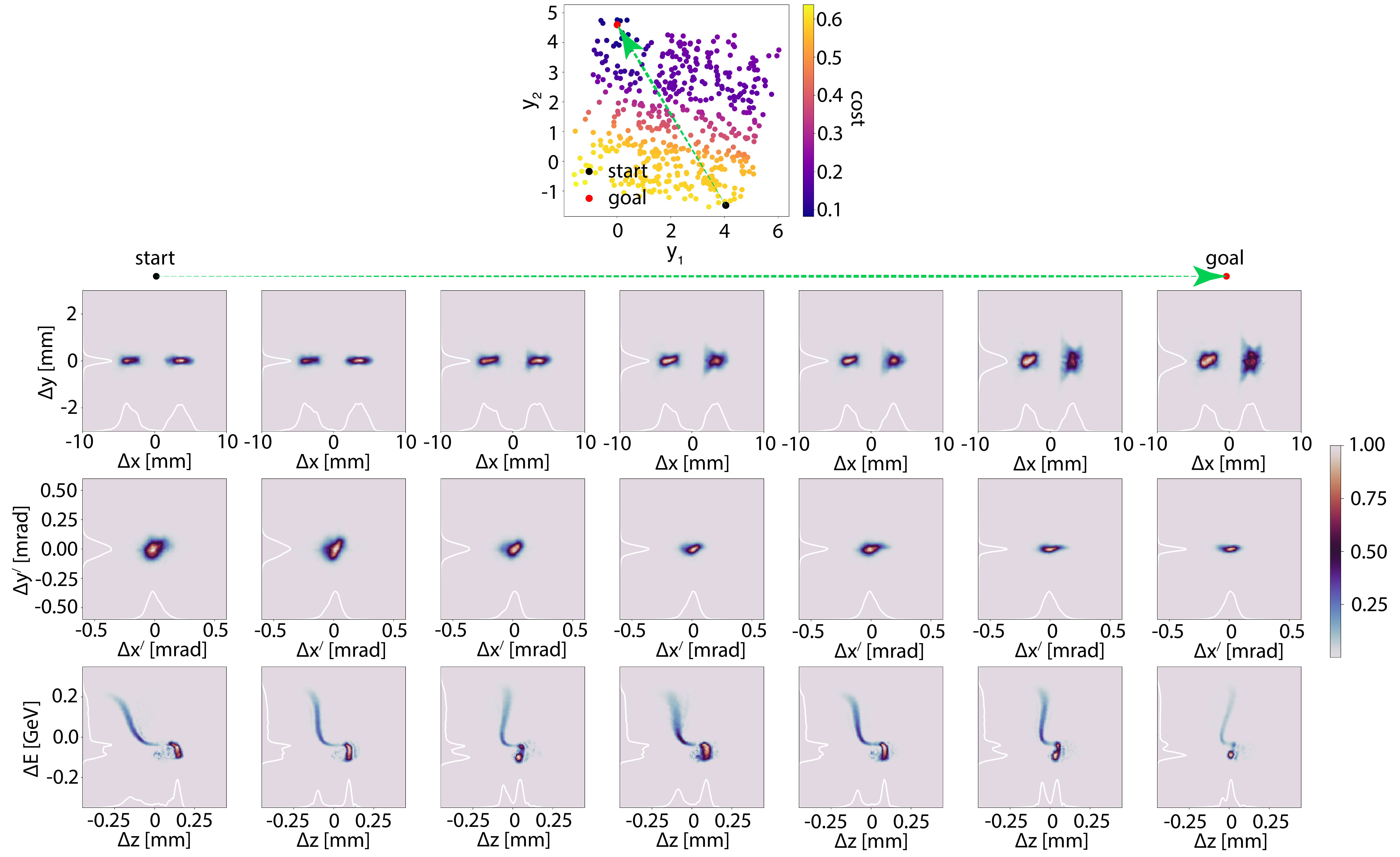}
    \caption{Moving in the 2D latent space from the start to the goal point. Seven equidistant $\rho_{13}(x,y)$ and $\rho_{14}(x,y')$ 2D phase space projections are shown at the location of TCAV1 and the $rho_{56}(z,E)$ 2D phase space projection is shown at the interaction point (IP) location. The encoder-decoder CNN has naturally organized the various distributions in the 2D latent space.}
    \label{fig:2D_LS}
\end{figure*}

Figure \ref{fig:2D_LS} shows the results of training the encoder-decoder CNN with a 2D latent space layer. The figure shows the 2D latent space positions $(y_1,y_2)$ of 500 test data sets which the CNN was not shown during training colored by the difference between the $\rho_{56}(z,E)$ phase space projection of the beam at the first TCAV1 as shown in the schematic of FACET-II defined by the equation: 
\begin{equation}
	\mathrm{cost} = \int_z\int_E \left |  \hat{\rho}_{56}(z,E) - \rho_{56}(z,E) \right | dEdz.
\end{equation}
Two points at opposite positions in the latent space are shown in red and black. As we move from the start to the goal point we see the evolution of the $\rho_{13}(x,y)$ and $\rho_{14}(x,y')$ projections at the location of TCAV1 and the $\rho_{56}(z,E)$ projection at the interaction point (IP) location. We see that the encoder section of the CNN has sorted the beam in the latent space so that as we move along the green dashed line the transverse beam size is increasing, the beam divergence is flattening, and the bunch-to-bunch separation is squeezed down to a single bunch.

\subsection{Latent Space Tuning for HiRES}
\label{sec:part}

\begin{figure*}[!tbh]
    \centering
    \includegraphics[width=0.6\textwidth]{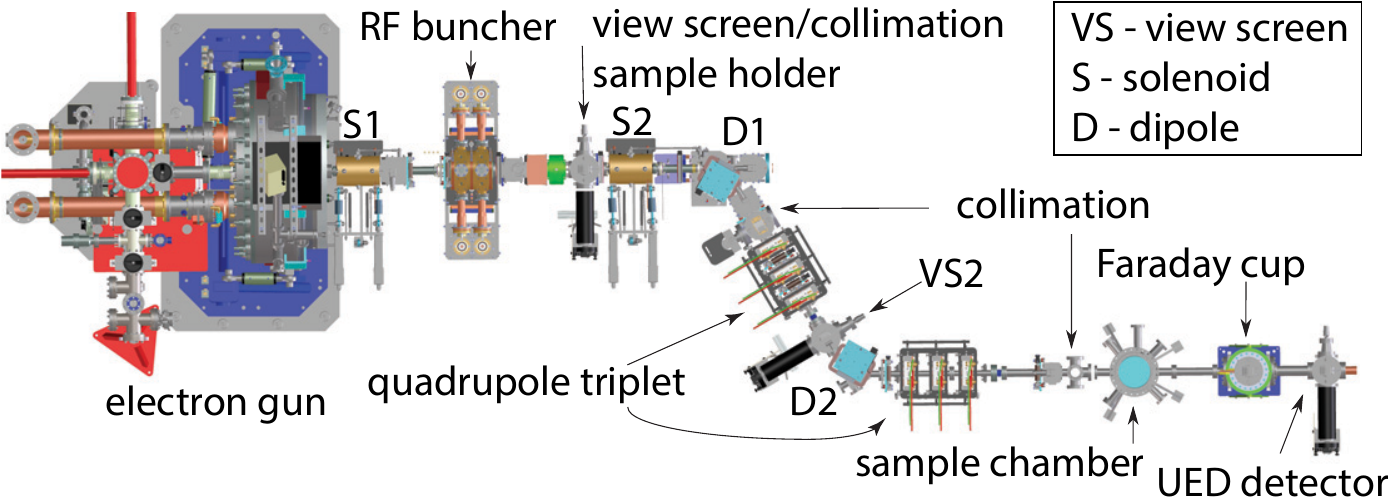}
    \includegraphics[width=0.9\textwidth]{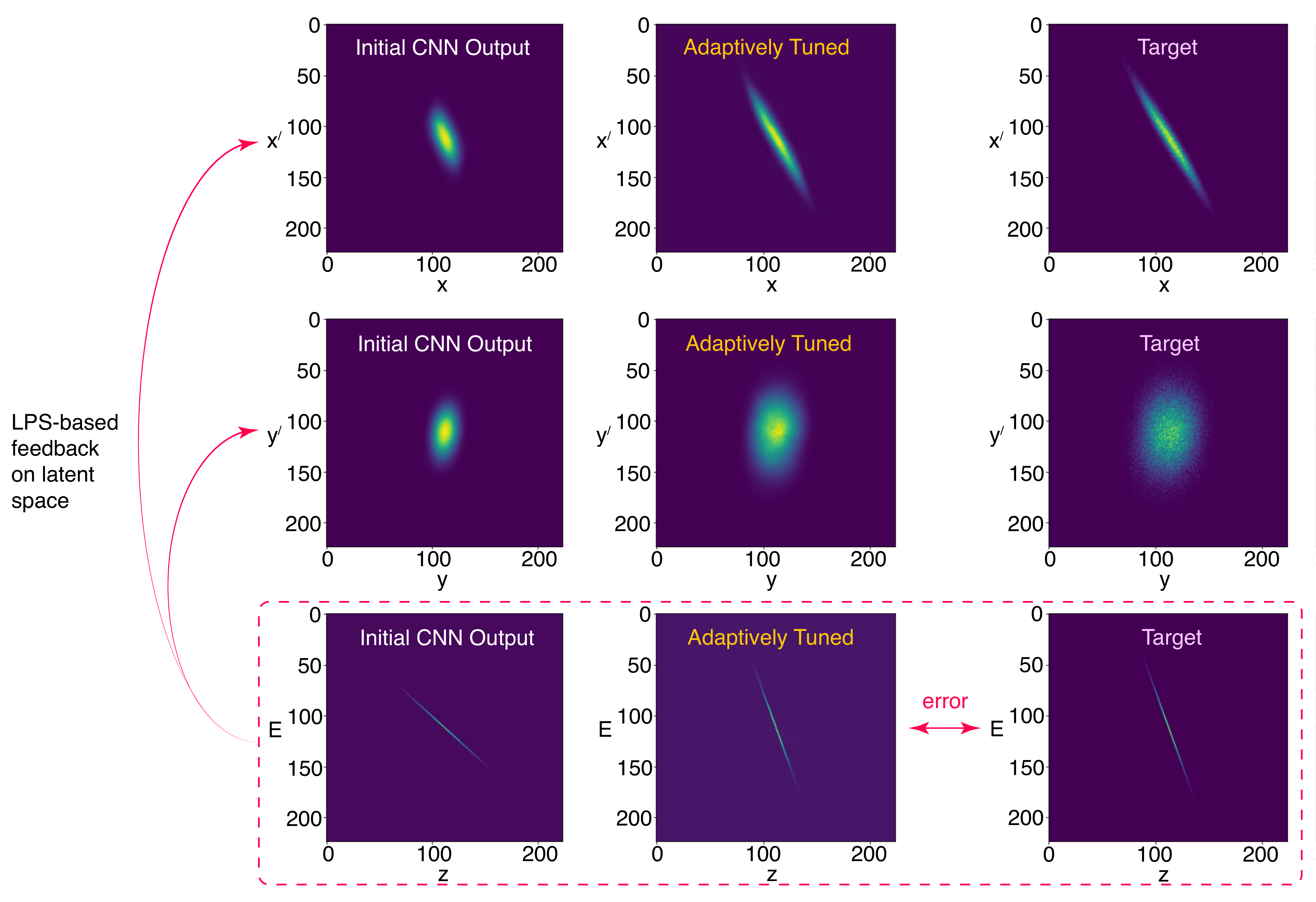}
    \caption{Predicting the transverse phase space $(x,x')$, $(y,y')$ based on LPS measurements $(z,E)$ at HiRES.}
    \label{fig:HiRES_AML}
\end{figure*}

It was first demonstrated in \cite{ref:LAML2} that an adaptive latent space tuning approach was applied at the HiRES ultra fast electron diffraction compact accelerator whose schematic is shown in Figure \ref{fig:HiRES_AML}. At HiRES the input distribution used was the $(x,y)$ phase space projection at the electron gun and the accelerator parameters included magnet strengths and beam energy. The latent space tuning approach enabled measurements of the $(z,E)$ longitudinal phase space to be mapped to the transverse phase space $(x,x')$, $(y,y')$ of the beam as shown in Figure \ref{fig:HiRES_AML}.

\subsection{Latent Space Tuning for FACET-II}
\label{sec:part}

For application to FACET-II, the $\mathbf{X}_I$ input in Figure \ref{fig:L_aml} is a 128$\times$128 pixel 2D $\rho_{13}(x,y)$ distribution of the $(x,y)$ phase space projection of the beam entering the particle accelerator and $\mathbf{x}_v$ is a vector of 7 accelerator and beam parameters which are the RF phase and amplitude settings, transverse beam offsets, and beam charge variation. The 16391 dimensional inputs $(\mathbf{X}_I,\mathbf{x}_v)$ are then combined into a single low-dimensional vector $\mathbf{y}_L$ of dimension between 2 and 100 before the generative half of the encoder-decoder CNN builds back up to a high dimensional output $\mathbf{\hat{Y}}$ which is a 128$\times$128$\times$75 output with the 75 channels representing a set of 15 2D projections of the 6D phase space: $(x,y)$, $(x,z)$, $(x,x')$, $(x,y')$, $(x,E)$, $(x',y)$, $(x',z)$, $(x',y')$, $(x',E)$, $(y,z)$, $(y,y')$, $(y,E)$, $(y',z)$, $(y',E)$, $(z,E)$ at 5 different accelerator locations in FACET-II as shown in Figure \ref{fig:75D}. 

%
\subsection{2D Latent Space}
%

\begin{figure*}[htbp]
\centering 
\includegraphics[width=1.0\textwidth]{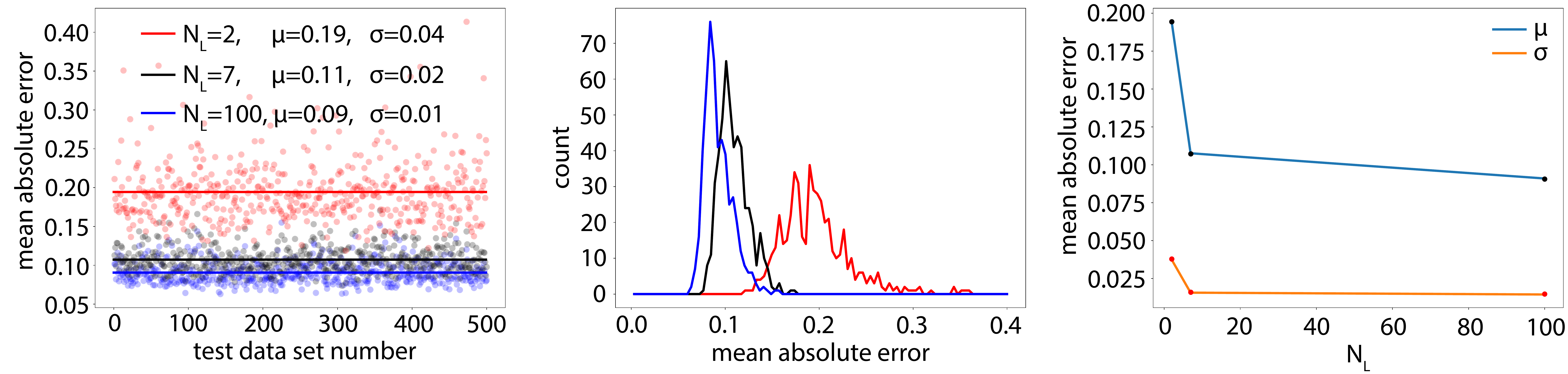}
\includegraphics[width=1.0\textwidth]{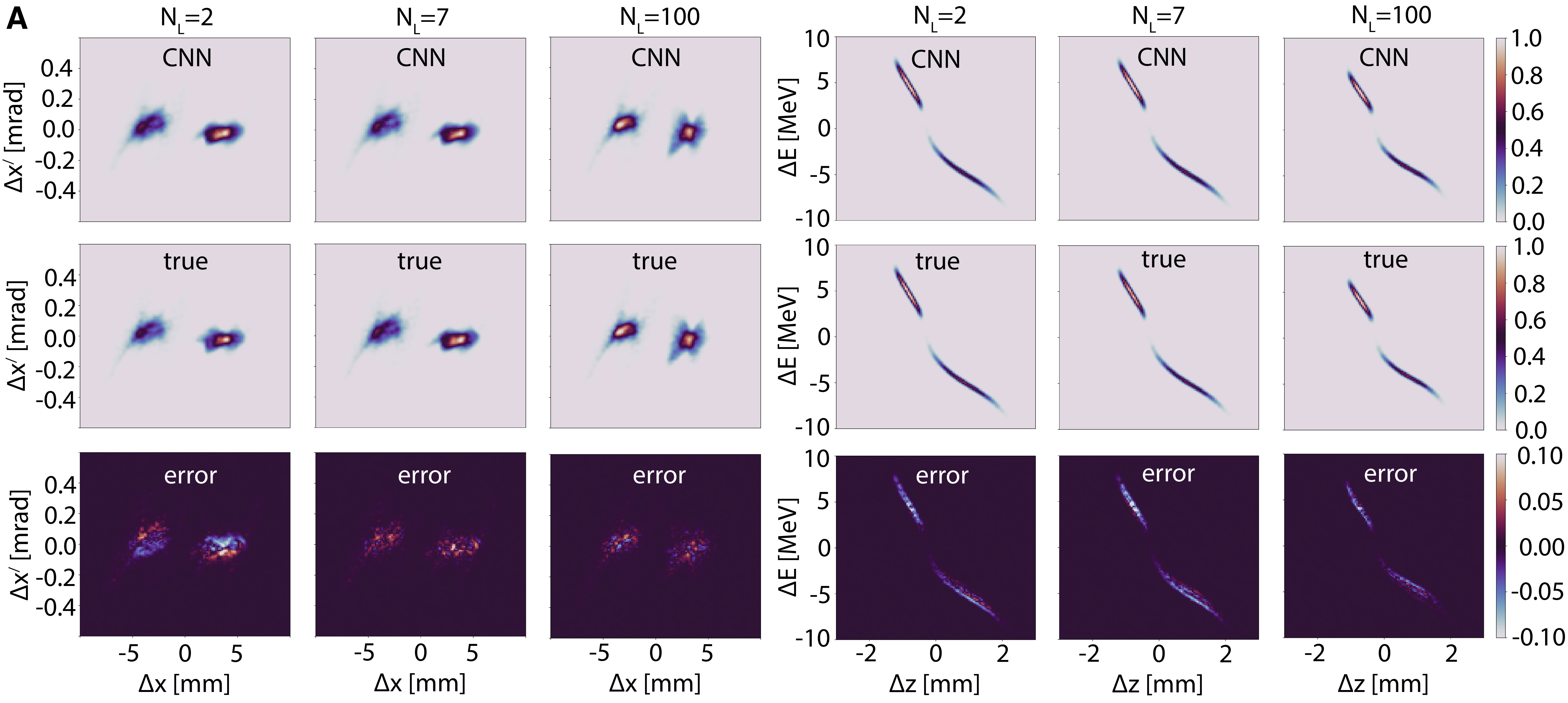}
\includegraphics[width=1.0\textwidth]{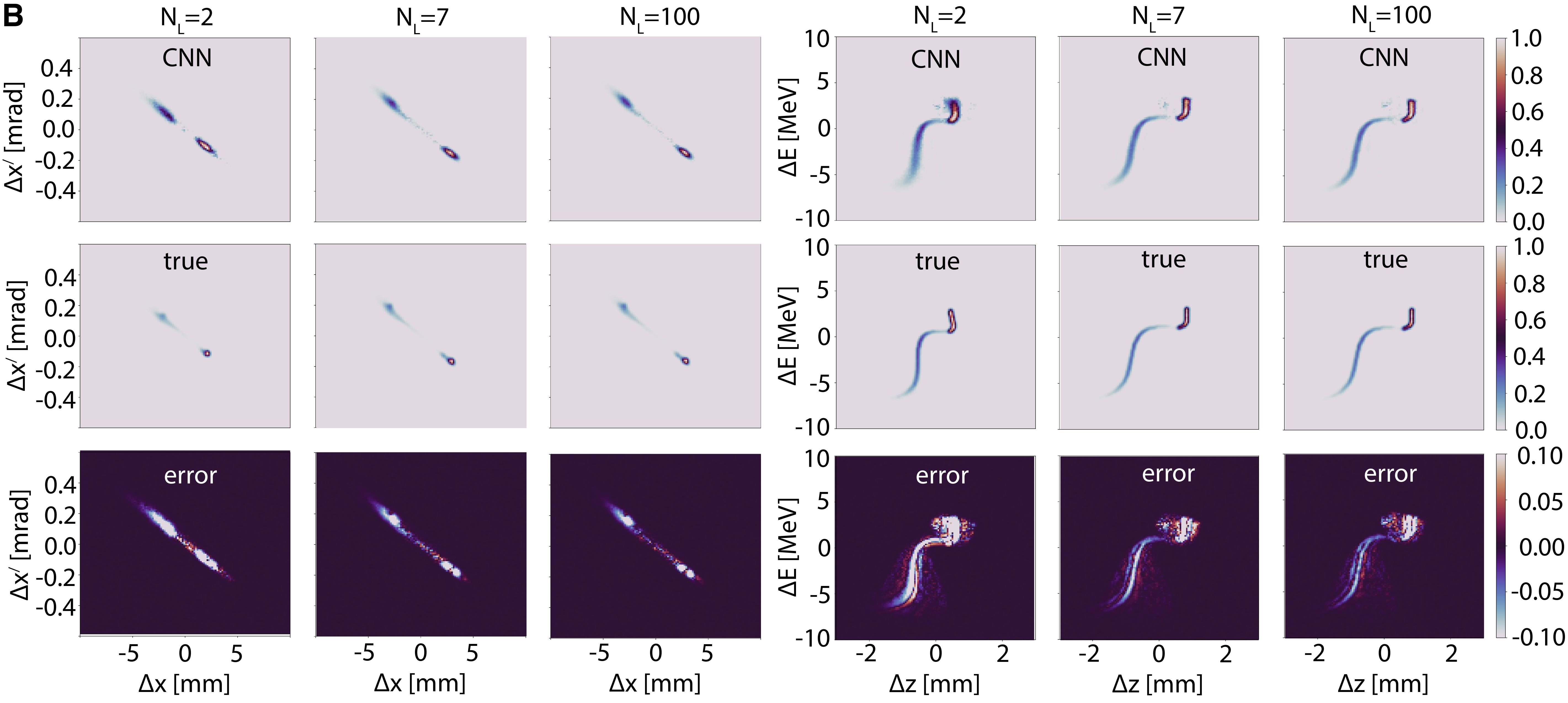}
\caption{\label{fig:2_7_10_Pred} Error statistics for 500 test data sets when using networks with latent space dimensions 2, 7, and 100 (Top). Comparison of prediction error for 2 projections of phase space for samples with approximately average error values of each CNN at the first TCAV (A) and at the interaction point (B).}
\end{figure*}

The CNN was trained with twenty thousand randomly generated input-output pairs, over a wide range. By forcing the CNN to simultaneously predict all 15 2D distribution projections which must satisfy all of the physics constraints of both general relativistic electrodynamics as well as the specific constraints of the FACET-II setup, the correlations between the various phase spaces at FACET-II were learned. Various latent space dimensions were tested including 2, 7, and 100, and the statistics of test-set errors are shown in Figure \ref{fig:2_7_10_Pred}. Clearly as the number of latent space dimensions is increased the accuracy of the CNN grows because the network size is bigger and the nonlinear representation is more general. However it is clear that after a dimension of 7 is reached the increase in accuracy is only incremental, which is understood by the fact that only 7 accelerator parameters were changed to create the training and test data for this demonstration.

The adaptive approach at FACET-II was to use an incorrect random (or it could just be the mean of all available measurements) guess for the input beam distribution and then iteratively adjust the low dimensional latent space to minimize a cost function which measured the difference between the predicted LPS $\hat{\rho}_{56}(z,E)$ and the correct LPS $\rho_{56}(z,E)$ according to several candidate cost functions defined as weighted combinations of:
\begin{eqnarray}
    C_{T,j} &=& \underbrace{\int_z\int_E \left |  \hat{\rho}_{56}(z,E) - \rho_{56}(z,E) \right | dEdz}_{\mathrm{at \ TCAVN}}, \quad N \in \left \{1,2,3 \right \}, \nonumber \\
    C_{B,j} &=& \underbrace{\int_x \left |  \hat{\rho}_{x}(E) - \rho_{x}(E) \right | dE}_{\mathrm{at \ BCN}}, \quad N \in \left \{ 11,14,20 \right \}, \quad \rho_x(E) = \int_{E} \rho_{16}(x,E)dE, \nonumber
\end{eqnarray}
where $C_{T,j}$ are costs based on TCAV measurements at the three TCAV locations considered for FACET-II and the $C_{B,j}$ are costs based on non-invasive energy spread spectra measurements which could possibly be collected at each of the three bunch compressors.

%
\subsection{Latent Space-Informed Diagnostics}
%

\begin{figure*}[htbp]
\centering 
\includegraphics[width=1.0\textwidth]{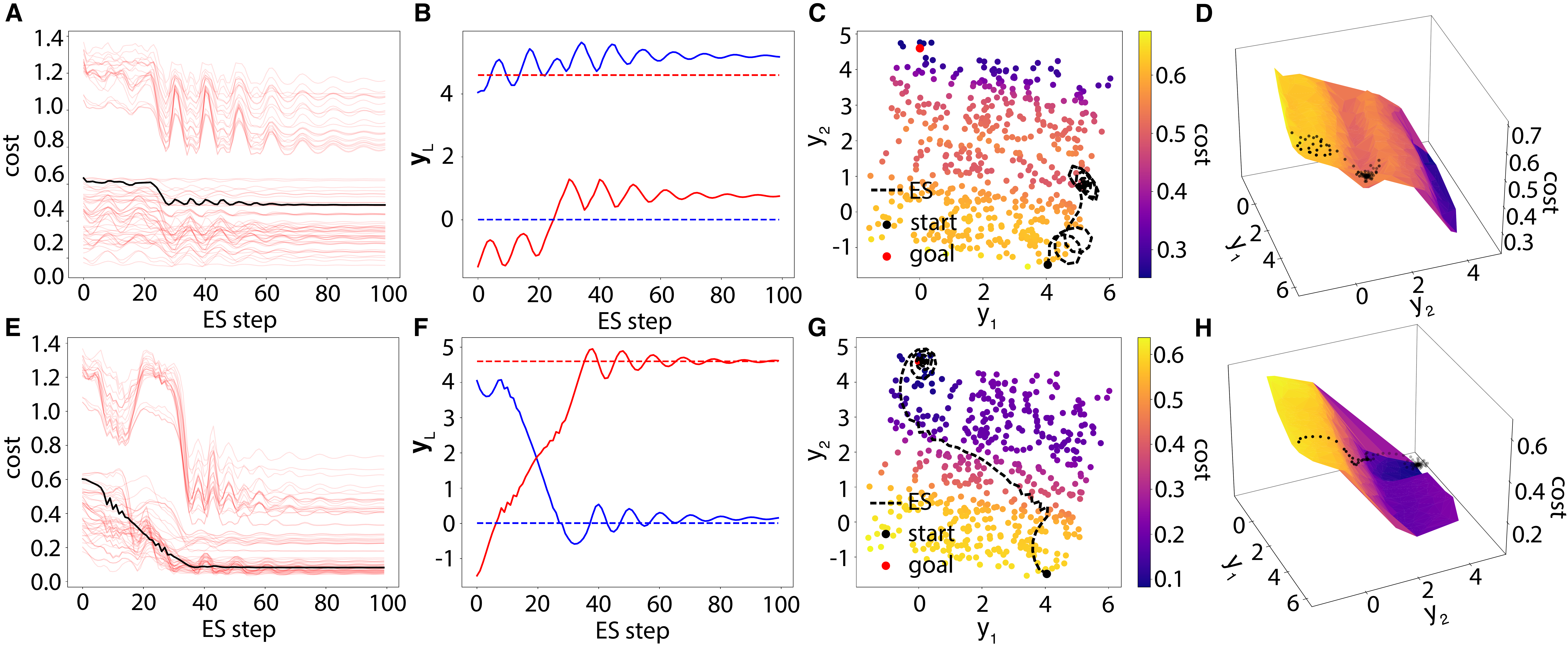}
\caption{\label{fig:ES_2D} Local and global min ES 2D.}
\end{figure*}

For the 2D latent space CNN a pair of adaptive tuning results are shown in Figure \ref{fig:ES_2D}. In Figure \ref{fig:ES_2D}A-D, the cost being adaptively minimized is the sum $\sum_{j=1}^{3}w_jC_{T,j}$, and we see that the resulting cost function has a local minimum near the starting point in the 2D latent space, as shown in more detail in \ref{fig:ES_2D}D. The result is that the adaptive scheme becomes stuck in this local minimum and is unable to reach the target. This counter-intuitive result can be understood by looking at Figure \ref{fig:2D_LS} where the IP $\hat{\rho}_{56}(z,E)$ phase space is shown, which is almost identical to the TCAV3 LPS. We can see that the way the CNN has sorted the 2D latent space, moving along the green dashed line from the starting to the goal position the bunch length is initially compressed and then spread out again before finally being compressed to a single bunch. This oscillating variation caused by including the TCAV3 measurement $\hat{\rho}_{56}(z,E)$ creates the local minimum in the cost function and a similar result happened with TCAV2. If we instead define the cost function to simply include only the TCAV1 LPS measurement, then we get the result shown in Figure \ref{fig:ES_2D}E-H, where the local minimum has disappeared and so the adaptive optimizer is able to make its way across the entire range of the latent space in order to adaptively match the unknown beam. 

This result is somewhat counter intuitive: using less information gives better convergence. One explanation is that FACET-II is such a large and complex machine that a given final beam phase space distribution can be created from many different initial distributions in a non-unique way based on all of the accelerator settings between the start and end of the machine. With this in mind it is understandable that the earlier diagnostics give a better chance of uniquely distinguishing between phase space distributions. This analysis shows that if a limited number of diagnostics must be chosen, such as placing a single TCAV at some location of an accelerator, it may be beneficial to place it earlier, as the measurements can be used to uniquely identify different beam characteristics. However there is of course an obvious downside to placing a diagnostic such as a TCAV too early in an accelerator configuration as one of the most useful applications of a TCAV is to measure the influence of collective effects such as space charge forces and CSR which grow in intensity as the beam reaches higher power and is compressed. Furthermore, although Figure \ref{fig:2_7_10_Pred} shows that the lower dimensional latent space results in less accurate results, Figure \ref{fig:ES_2D} shows the benefit of a low-dimensional latent space in that convergence is extremely fast, taking as few as 40 steps.

%
\subsection{$ND, N>2$ latent space}
%

\begin{figure*}[htbp]
\centering 
\includegraphics[width=1.0\textwidth]{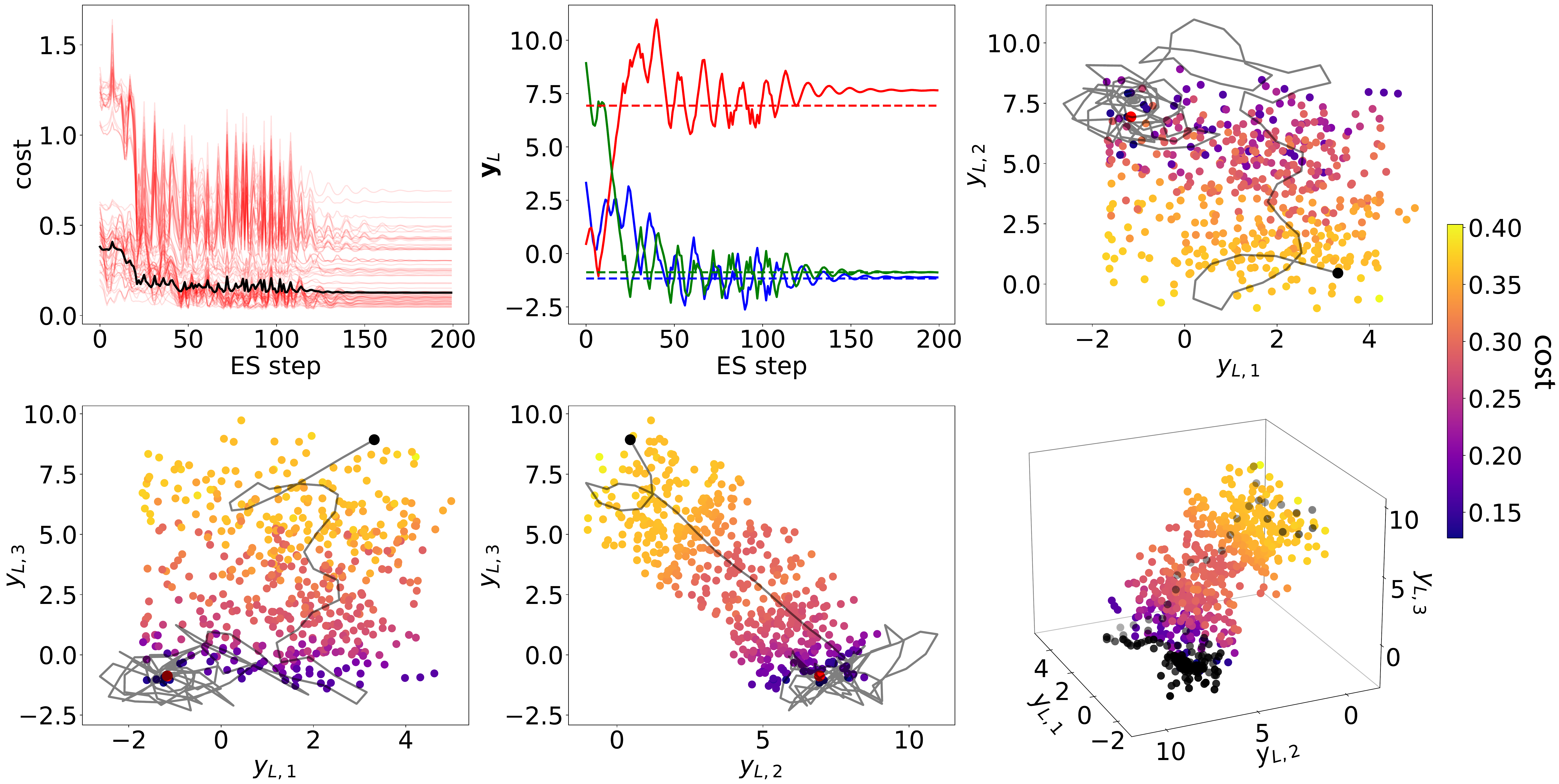}
\caption{\label{fig:ES_3D} Convergence in the 3D latent space is shown, which takes approximately 150 steps. We show all 3 2D projections as well as the evolution of the trajectory (black) in the full 3D latent space with the underlying test set-based latent space colored by the value of the cost function.}
\end{figure*}

\begin{figure*}[htbp]
\centering 
\includegraphics[width=1.0\textwidth]{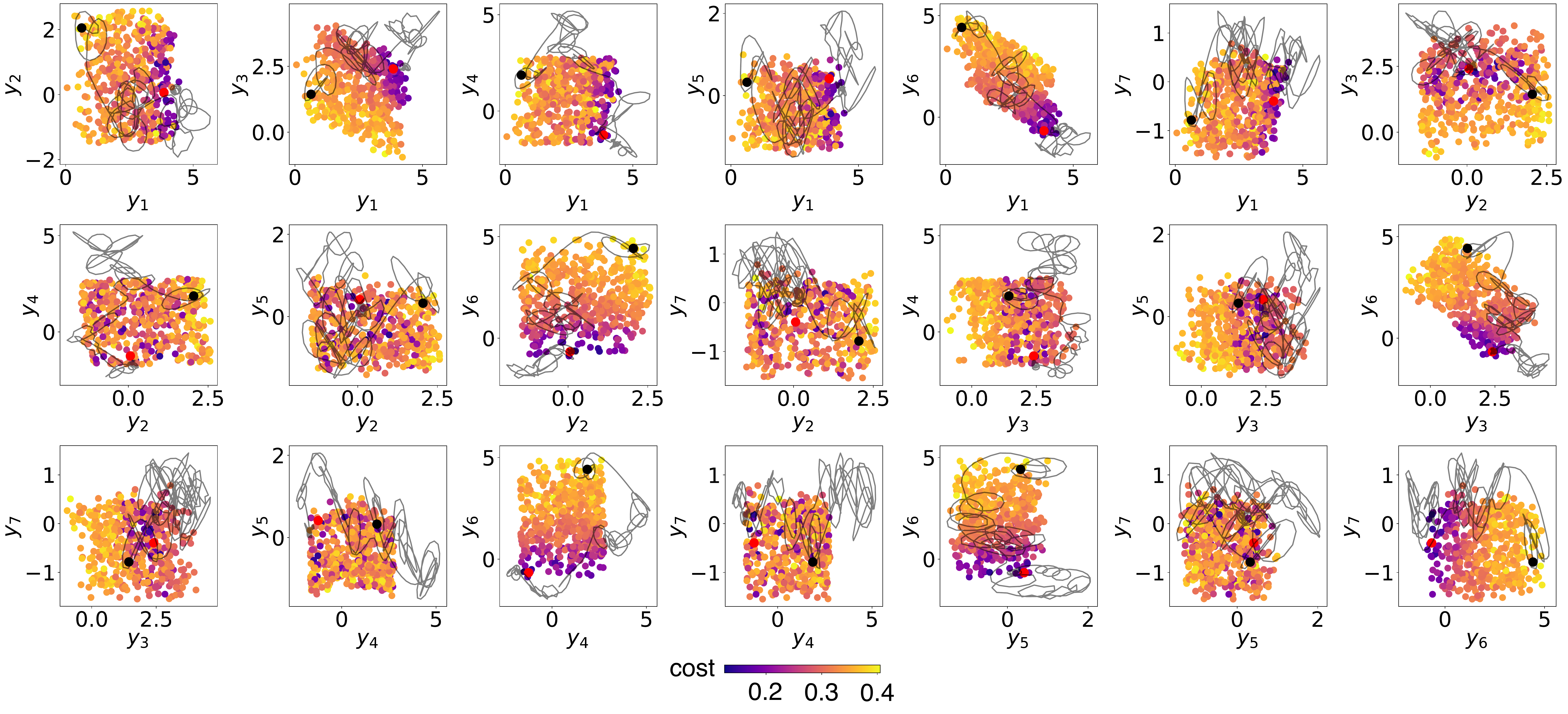}
\caption{\label{fig:ES_7D_global} The adaptive feedback trajectory in 7D latent space is shown (black) on top of all 21 projections of 2D views of the 7D latent space, which is colored by the value of the cost function being minimized.}
\end{figure*}

\begin{figure*}[htbp]
\centering 
\includegraphics[width=1.0\textwidth]{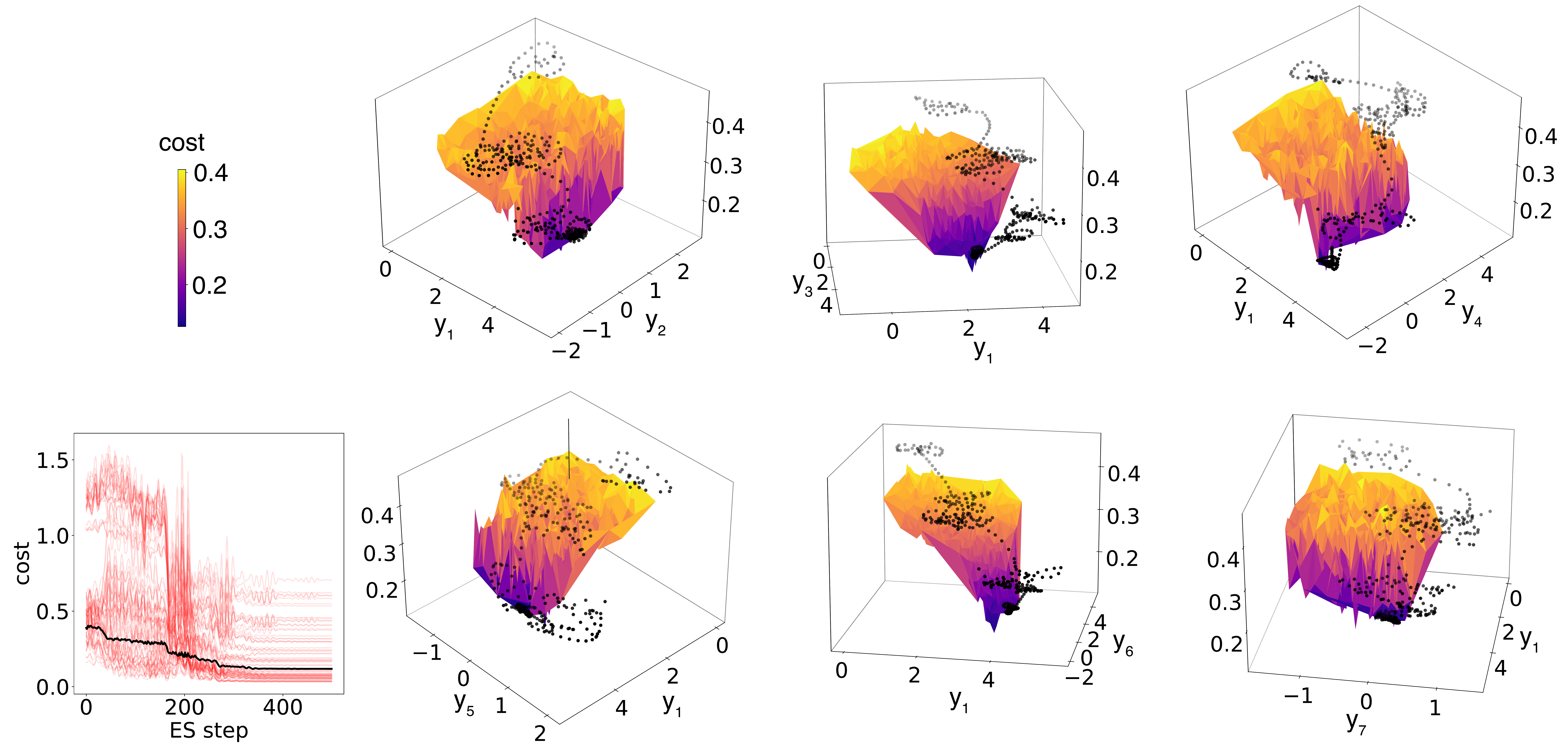}
\caption{\label{fig:ES_7D_global_3D} Several 3D projections $(y_1,y_n)$ for $n\in\{2,3,4,5,6,7\}$ of convergence within the 7D latent space are shown with the adaptively tuned trajectory shown as black dots lifted slightly above the surface of the cost function. The cost convergence is also shown and seen to take approximately 400 steps to converge.}
\end{figure*}

Higher dimensional latent spaces result in more accurate CNN predictions, they also make it less likely to get stuck in a local minimum as it is easier to avoid local minima in a higher dimensional space. Figure \ref{fig:ES_3D} shows the results of tuning in a 3D latent space when using a cost which includes all of the TCAV and BC terms. The global minimum is reached by taking a wider path that avoids local minima. Figures \ref{fig:ES_7D_global},\ref{fig:ES_7D_global_3D} show tconvergence in a 7D latent space. Again the cost function includes all terms and the global minimum is reached with more space to avoid local minima.

\begin{figure*}[htbp]
\centering 
\includegraphics[width=1.0\textwidth]{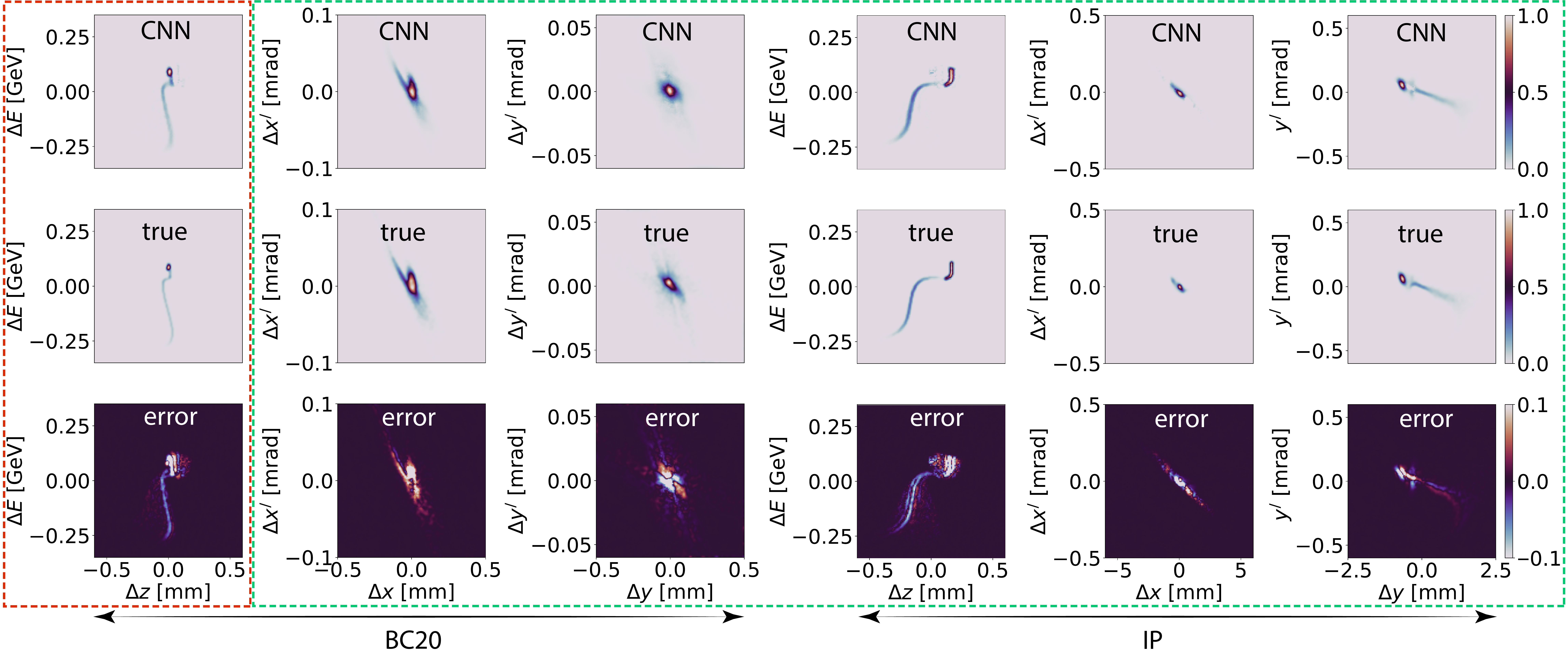}
\caption{\label{fig:NL7_Pred} Predictions of the 7D latent space model of the phase space at BC20 and at the IP. The red dashed box shows a LPS diagnostic that was used as part of the cost function while the other 2D phase space projections in the green dashed box were unseen by the CNN.}
\end{figure*}

However, a major downside of increasing the latent space dimension is clearly seen in the growing number of steps required for convergence to take place. In practice there is a tradeoff between simplicity and accuracy/generality that must be made when choosing the latent space dimension for a given problem. Figure \ref{fig:NL7_Pred} shows the results of the 7D latent space convergence in Figures \ref{fig:ES_7D_global}, \ref{fig:ES_7D_global_3D}. Figure \ref{fig:NL7_Pred} demonstrates that despite using only the projections measured for the cost functions $C_{T,j}$ and $C_{B,J}$ as defined above, the other 2D projections are also accurately predicted. We emphasize that the CNN is actually predicting all 75 projections which are 5 sets of 15 projections at 5 locations in FACET-II, but to save space we just show a few of the projections. 

Furthermore, we emphasize that both the input beam distributions and the accelerator parameters in the test set were unseen during the CNN's training process and the CNN is now predicting the correct 2D phase space distributions without needing access to accurate input beam distributions or to accelerator parameters. This is the major strength of this approach which makes it robust to time-varying systems for which invasive input beam distributions are no longer available after an initial dedicated training period which might take place during a dedicated beam development time.

%
\section{Conclusions}
%

An adaptive latent space tuning method has been developed and demonstrated on FACET-II and HiRES based on encoder-decoder style CNNs whose generative halves predict various 2D phase space projections of charged particle beams based only on a single 2D phase space slice measurement such as the LPS provided by TCAVs. Unlike traditional ML-based methods this is robust to unknown time-varying input beam distributions which are represented by the low dimensional latent space of the CNN which is quickly and automatically adjusted based on model-independent feedback.

%
\section{Acknowledgements} 
%

This work was funded by the Los Alamos National Laboratory LDRD Program project 20210160DR as well as the Department of Energy Office of Science Basic Energy Sciences High Energy Physics Program on Data, Artificial Intelligence, and Machine Learning at DOE Scientific User Facilities, LAB 20-2261, project "Leveraging ML/AI techniques to enable a breakthrough in ultra-short-bunch paradigm."


\begin{thebibliography}{99}

	
	\bibitem{ref:LCLS}
    	A. A. Lutman, et al. ``Fresh-slice multicolour X-ray free-electron lasers." Nature Photonics 10.11 (2016): pp. 745-750.
	
	\bibitem{ref:SwissFEL} 
    	A. Malyzhenkov, et al. ``Single-and two-color attosecond hard x-ray free-electron laser pulses with nonlinear compression." Physical Review Research 2.4 (2020): 042018.
	
	\bibitem{ref:EuXFEL}
		W. Decking, et al. ``A MHz-repetition-rate hard X-ray free-electron laser driven by a superconducting linear accelerator." Nature Photonics 14.6 (2020): 391-397.
		
	
	\bibitem{ref:FACET2}
		V. Yakimenko, et al. ``FACET-II facility for advanced accelerator experimental tests." Physical Review Accelerators and Beams 22.10 (2019): 101301.
		
	\bibitem{ref:AWAKE}
    	E. Adli, et al. ``Acceleration of electrons in the plasma wakefield of a proton bunch.” Nature 561, 363–367 (2018).
    
    
    	\bibitem{ref:FACET_LPS_1}
        C. Emma, et al. ``Machine learning-based longitudinal phase space prediction of particle accelerators." Physical Review Accelerators and Beams 21.11 (2018): 112802.
        
    	\bibitem{ref:FACET_LPS_2}
        A. Hanuka, et al. ``Accurate and confident prediction of electron beam longitudinal properties using spectral virtual diagnostics." Scientific Reports 11.1 (2021): 1-10.
        
   	\bibitem{ref:LCLS2_Injector} L. Gupta, et al. "Improving Surrogate Model Accuracy for the LCLS-II Injector Frontend Using Convolutional Neural Networks and Transfer Learning." arXiv preprint arXiv:2103.07540, 2021.
   
   	\bibitem{ref:deep_Q} S. Hirlaender and N. Bruchon. ``Model-free and Bayesian Ensembling Model-based Deep Reinforcement Learning for Particle Accelerator Control Demonstrated on the FERMI FEL." arXiv preprint arXiv:2012.09737, 2020.
	
	\bibitem{ref:PE} A. Adelmann, ``On nonintrusive uncertainty quantification and surrogate model construction in particle accelerator modeling." SIAM/ASA Journal on Uncertainty Quantification, 7, 383–416, 2019.
	
	\bibitem{ref:UQ} O. Convery, L. Smith, Y. Gal, and A. Hanuka. ``Uncertainty Quantification for Virtual Diagnostic of Particle Accelerators." arXiv preprint arXiv:2105.04654, 2021.

	\bibitem{ref:TS} S. Li, et al. ``A Novel Approach for Classification and Forecasting of Time Series in Particle Accelerators." {\it Information}, 12.3, 121, 2021.
	
	\bibitem{ref:MO1} A. Edelen, et al. ``Machine learning for orders of magnitude speedup in multiobjective optimization of particle accelerator systems." {\it Physical Review Accelerators and Beams} 23.4, 044601, 2020.

	\bibitem{ref:MO2} M. Kranjcevic, B. Riemann, A. Adelmann, and A. Streun. ``Multiobjective optimization of the dynamic aperture using surrogate models based on artificial neural networks." {\it Physical Review Accelerators and Beams}, 24(1), 014601, 2021.
   
   	\bibitem{ref:GP_1} M. McIntire, et al. ``Bayesian optimization of FEL performance at LCLS." Proceedings of the 7th International Particle Accelerator Conference, 2016.

	\bibitem{ref:GP_2} Y. Li, R. Rainer, W. Cheng. ``Bayesian approach for linear optics correction." {\it Physical Review Accelerators and Beams}, 22, 012804, 2019.

	\bibitem{ref:GP_3} Y. Hao, et al. ``Reconstruction of Storage Ring's Linear Optics with Bayesian Inference." arXiv preprint arXiv:1902.11157, 2019.

 	\bibitem{ref:GP_4} J. Duris, et al. ``Bayesian optimization of a free-electron laser." {\it Physical review letters} 124.12, 124801, 2020.
   
   	\bibitem{ref:GP_5} A. Hanuka, et al. ``Online tuning and light source control using a physics-informed Gaussian process Adi." arXiv preprint arXiv:1911.01538, 2019.

	\bibitem{ref:GP_6} Y. Li, Y. Hao, W. Cheng, and R. Rainer. ``Analysis of beam position monitor requirements with Bayesian Gaussian regression." arXiv preprint arXiv:1904.05683, 2019.

	\bibitem{ref:GP_7} R. J. Shalloo, et al. ``Automation and control of laser wakefield accelerators using Bayesian optimization." {\it Nature communications} 11.1, 1-8, 2020.
	
	\bibitem{ref:GP_8} R. Roussel, A. Hanuka, and A. Edelen. ``Multiobjective Bayesian optimization for online accelerator tuning." {\it Physical Review Accelerators and Beams}, 24.6 062801, 2021.
   
   	\bibitem{ref:ML_CERN1} E. Fol, J. C. de Portugal, G. Franchetti, R. Tomas. ``Optics corrections using machine learning in the LHC." Proceedings of the 2019 International Particle Accelerator Conference, Melbourne, Australia, 2019.

	\bibitem{ref:ML_CERN2} E. Fol, et al. ``Unsupervised Machine Learning for Detection of Faulty Beam Position Monitors." Proc. 10th Int. Particle Accelerator Conf.(IPAC’19), Melbourne, Australia, Vol. 2668, 2019.

	\bibitem{ref:ML_CERN3} P. Arpaia et al. ``Machine learning for beam dynamics studies at the CERN Large Hadron Collider." {\it Nuclear Instruments and Methods in Physics Research Section A: Accelerators, Spectrometers, Detectors and Associated Equipment}, 985, 164652, 2021.
	
	\bibitem{ref:ES}
        A. Scheinker and M. Krsti\'c. ``Minimum-seeking for CLFs: Universal semiglobally stabilizing feedback under unknown control directions." IEEE Transactions on Automatic Control 58.5 (2012): 1107-1122.
        
        
        \bibitem{ref:ES_ML}
        A. Scheinker, et al. ``Demonstration of model-independent control of the longitudinal phase space of electron beams in the linac-coherent light source with femtosecond resolution." Physical review letters 121.4 (2018): 044801.
        
    	\bibitem{ref:ES_AML} 
    	A. Scheinker. ``Adaptive Machine Learning for Robust Diagnostics and Control of Time-Varying Particle Accelerator Components and Beams." {\it Information}, 12.4, 161, 2021.
	
	\bibitem{ref:LAML1} A. Scheinker, F. Cropp, S. Paiagua, and D. Filippetto. ``Adaptive deep learning for time-varying systems with hidden parameters: Predicting changing input beam distributions of compact particle accelerators." arXiv preprint arXiv:2102.10510, 2021.

	\bibitem{ref:LAML2} A. Scheinker, F. Cropp, S. Paiagua, and D. Filippetto. ``Adaptive Latent Space Tuning for Non-Stationary Distributions." arXiv preprint arXiv:2105.03584, 2021.
        

	\bibitem{ref:Khalil}
    	H. K. Khalil. {\it Nonlinear Systems}. Prentice Hall, 2002.
    
    	\bibitem{ref:Ioannou1}
    	P. Ioannou and B. Fidan. {\it Adaptive control tutorial.} Society for Industrial and Applied Mathematics, 2006.
	
    	\bibitem{ref:Ioannou2}
    	P. A. Ioannou and J. Sun. {\it Robust adaptive control}. Courier Corporation, 2012.

        
        \bibitem{ref:ES2}
        A. Scheinker. "Simultaneous stabilization and optimization of unknown, time-varying systems." 2013 American Control Conference. IEEE, 2013.
        
        \bibitem{ref:ES3} 
        A. Scheinker and D. Scheinker. ``Bounded extremum seeking with discontinuous dithers." {\it Automatica}, 69, 250-257, 2016.
        
        \bibitem{ref:ES4} A. Scheinker. ``Model independent beam tuning." Proceedings of the 2013 International Particle Accelerator Conference, Shanghai, China. 2013.

	\bibitem{ref:ES5} A. Scheinker and M. Krstic. ``Extremum seeking with bounded update rates." {\it Systems \& Control Letters}, 63, 25-31, 2014.

	\bibitem{ref:krizhevsky2012imagenet} 
	A. Krizhevsky, I. Sutskever, and G. E. Hinton. ``Imagenet classification with deep convolutional neural networks." Advances in neural information processing systems 25 (2012): 1097-1105.

	\bibitem{ref:PCA}
        S. Wold, K. Esbensen, and P. Geladi. ``Principal component analysis." Chemometrics and intelligent laboratory systems 2.1-3 (1987): 37-52.
        
        
        \bibitem{ref:TCAV}
        C. Behrens, et al. ``Few-femtosecond time-resolved measurements of X-ray free-electron lasers." Nature communications 5.1 (2014): 1-7.

	\bibitem{ref:ES_FACET}
        A. Scheinker and S. Gessner. ``Adaptive method for electron bunch profile prediction." Physical Review Special Topics-Accelerators and Beams 18.10 (2015): 102801.
        
    	\bibitem{ref:ES_FACET_2}
        A. Scheinker, et al. ``Adaptive model tuning studies for non-invasive diagnostics and feedback control of plasma wakefield acceleration at FACET-II." Nuclear Instruments and Methods in Physics Research Section A: Accelerators, Spectrometers, Detectors and Associated Equipment 967 (2020): 163902.
        
        
    	




\end{thebibliography}
\end{document}